\documentclass{article}

\usepackage{graphics}
\usepackage{graphicx}
\usepackage{wrapfig}
\usepackage{hyperref}
\usepackage{amsfonts}
\usepackage{algorithm,algorithmicx}
\usepackage{algpseudocode}
\usepackage{microtype}      
\usepackage{pdfpages}
\usepackage{multicol}
\usepackage{bookmark}
\usepackage{amsmath}
\usepackage{amssymb}
\usepackage{float}
\usepackage{amsthm}
\usepackage[toc,page]{appendix}

\usepackage[accepted]{icml2021}

\icmltitlerunning{Energy-based Surprise Minimization for Multi-Agent Value Factorization}

\begin{document}

\twocolumn[
\icmltitle{Energy-based Surprise Minimization for Multi-Agent Value Factorization}



\icmlsetsymbol{equal}{*}

\begin{icmlauthorlist}
\icmlauthor{Karush Suri}{to,goo}
\icmlauthor{Xiao Qi Shi}{ed}
\icmlauthor{Konstantinos Plataniotis}{to}
\icmlauthor{Yuri Lawryshyn}{to,goo}

\end{icmlauthorlist}

\icmlaffiliation{to}{Department of Electrical \& Computer Engineering, University of Toronto, Canada.}
\icmlaffiliation{goo}{Center for Management of Technology \& Entrepreneurship (CMTE)}
\icmlaffiliation{ed}{RBC Capital Markets}

\icmlcorrespondingauthor{Karush Suri}{karush.suri@mail.utoronto.ca}

\icmlkeywords{ESAC, mutation, AMT, policy}

\vskip 0.3in
]

\newtheorem{innercustomgeneric}{\customgenericname}
\providecommand{\customgenericname}{}
\newcommand{\newcustomtheorem}[2]{%
  \newenvironment{#1}[1]
  {%
   \renewcommand\customgenericname{#2}%
   \renewcommand\theinnercustomgeneric{##1}%
   \innercustomgeneric
  }
  {\endinnercustomgeneric}
}

\newcustomtheorem{customthm}{Theorem}
\newcustomtheorem{customlemma}{Lemma}



\printAffiliationsAndNotice{} 

\begin{abstract}
    Multi-Agent Reinforcement Learning (MARL) has demonstrated significant success in training decentralised policies in a centralised manner by making use of value factorization methods. However, addressing surprise across spurious states and approximation bias remain open problems for multi-agent settings. Towards this goal, we introduce the \textit{Energy-based MIXer (EMIX)}, an algorithm which minimizes surprise utilizing the energy across agents. Our contributions are threefold; (1) EMIX introduces a novel surprise minimization technique across multiple agents in the case of multi-agent partially-observable settings. (2) EMIX highlights a practical use of energy functions in MARL with theoretical guarantees and experiment validations of the energy operator. Lastly, (3) EMIX extends Maxmin Q-learning for addressing overestimation bias across agents in MARL. In a study of challenging StarCraft II micromanagement scenarios, EMIX demonstrates consistent stable performance for multi-agent surprise minimization. Moreover, our ablation study highlights the necessity of the energy-based scheme and the need for elimination of overestimation bias in MARL. Our implementation of EMIX can be found at \href{https://karush17.github.io/emix-web/}{karush17.github.io/emix-web/}.
\end{abstract}

\section{Introduction}
Reinforcement Learning (RL) has seen tremendous growth in applications such as arcade games \cite{atari}, board games \cite{go, shogi}, robot control tasks \cite{ddpg, ppo} and lately, real-time games \cite{SC2}. The rise of RL has led to an increasing interest in the study of multi-agent systems \cite{maddpg, alphastar}, commonly known as Multi-Agent Reinforcement Learning (MARL). In the case of partially observable settings, MARL enables the learning of policies with centralised training and decentralised control \cite{dec}. This has proven to be useful for exploiting value-based methods which are otherwise deamed to be sample-inefficient \cite{iql,coma}.

\indent Value Factorization \cite{vdn,qmix} is a common technique which enables the joint value function to be represented as a combination of individual value functions conditioned on states and actions. In the case of Value Decomposition Network (VDN) \cite{vdn}, a linear additive factorization is carried out whereas QMIX \cite{qmix} generalizes the factorization to a non-linear combination, hence improving the expressive power of centralised action-value functions. Furthermore, monotonicity constraints in QMIX enable scalability in the number of agents. On the other hand, factorization across multiple value functions leads to the aggregation of approximation bias \cite{doubleqlearning, deepdoubleqlearning} originating from overoptimistic estimations in action values \cite{td3,maxmin}. Moreover, value factorization methods yield state-based estimates and do not account for spurious changes in partially-observed observations, commonly referred to as surprise \cite{surprise}.

\indent \textit{Surprise minimization} \cite{smirl} is a recent phenomenon observed in the case of single-agent RL methods which deals with environments consisting of spurious states. In the case of model-based RL \cite{mbrl}, surprise minimization is used as an effective planning tool in the agent's model \cite{smirl} whereas in the case of model-free RL, surprise minimization is witnessed as an intrinsic motivation \cite{surprise,surpmodeling} or generalization problem \cite{gen}. On the other hand, utilization of surprise minimization towards MARL has been limited as a result of which agents remain unaware of drastic changes in the environment \cite{role}. Thus, surprise minimization in the case of multi-agent settings requires attention from a critical standpoint.

\indent We introduce the Energy-based MIXer (EMIX), an algorithm based on QMIX which minimizes surprise utilizing the energy across agents. Our contributions are threefold; (1) EMIX introduces a novel surprise minimization technique across multiple agents in the case of multi-agent partially-observable settings. (2) EMIX highlights a practical use of energy functions in MARL with theoretical guarantees and experiment validations of the energy operator. Lastly, (3) EMIX extends Maxmin Q-learning \cite{maxmin} addressing overestimation bias across agents in MARL. When evaluated on a range of challenging StarCraft II scenarios \cite{smac}, EMIX demonstrates consistently stable performance for multi-agent surprise minimization by improving the QMIX framework. Moroever, our ablation study highlights the necessity of our energy-based scheme and the need for elimination of overestimation bias in MARL.



\section{The Value Factorization Problem}
\subsection{Preliminaries}
We review the cooperative MARL setup. The problem is modeled as a Partially Observable Markov Decision Process (POMDP) \cite{rl} defined by the tuple $(\mathcal{S},\mathcal{A},r,N,P,Z,O,\gamma)$ where the state space $\mathcal{S}$ and action space $\mathcal{A}$ are discrete, $r: \mathcal{S} \times \mathcal{A} \rightarrow [r_{min},r_{max}]$ presents the reward observed by agents $a \in N$ where $N$ is the set of all agents, $P: \mathcal{S} \times \mathcal{S} \times \mathcal{A} \rightarrow [0,\infty)$ presents the unknown transition model consisting of the transition probability to the next state $s\textquotesingle \in \mathcal{S}$ given the current state $s \in \mathcal{S}$ and joint action $u \in \mathcal{A}$ at time step $t$ and $\gamma$ is the discount factor. We consider a partially observable setting in which each agent $n$ draws individual observations $z \in Z$ according to the observation function $O(s,u): \mathcal{S} \times \mathcal{A} \rightarrow Z$. We consider a joint policy $\pi_{\theta}({u|s})$ as a function of model parameters $\theta$. Standard RL defines the agent's objective to maximize the expected discounted reward $\mathbb{E}_{\pi_{\theta}}[\sum_{t=0}^{T}\gamma^{t}r(s_{t},u_{t})]$ as a function of the parameters $\theta$. The action-value function for an agent is represented as $Q(u,s;\theta) = \mathbb{E}_{\pi_{\theta}}[\sum_{t=1}^{T}\gamma^{t}r(s,u)|s=s_{t},u=u_{t}]$ which is the expected sum of payoffs obtained in state $s$ upon performing action $u$ by following the policy $\pi_{\theta}$. We denote the optimal policy $\pi_{\theta}^{*}$ such that $Q(u,s;\theta^{*}) \geq Q(u,s;\theta) \forall s \in S, u \in A$. In the case of multiple agents, the joint optimal policy can be expressed as the Nash Equilibrium \cite{nash} of the Stochastic Markov Game as $\pi^{*} = (\pi^{1,*},\pi^{2,*}, ...\pi^{N,*})$ such that $Q(u^{a},s;\theta^{*}) \geq Q(u^{a},s;\theta) \forall s \in S, u \in A, a \in N$. 

Q-Learning is an off-policy, model-free algorithm suitable for continuous and episodic tasks. The algorithm uses semi-gradient descent to minimize the Temporal Difference (TD) error: $\mathbb{L(\theta)} = \underset{b \sim R}{\mathbb{E}}[(y - Q(u,s;\theta))^{2}]$ where $y = r + \gamma \underset{u^{\textquotesingle}\in A}{\max} Q(u^{\textquotesingle},s^{\textquotesingle};\theta^{-})$ is the TD target consisting of $\theta^{-}$ as the target parameters and $b$ is the batch sampled from memory $R$.

\subsection{Surprise Minimization}
Despite the recent success of value-based methods \cite{a3c,rainbow} RL agents suffer from spurious state spaces and encounter sudden changes in trajectories. These anomalous transitions between consecutive states are termed as \textit{surprise} \cite{surprise}. Quantitatively, surprise can be inferred as a measure of deviation \cite{smirl,gen} among states encountered by the agent during its interaction with the environment. 

While exploring \cite{curiosity,exploration} the environment, agents tend to have higher deviation among states which is gradually reduced by gaining a significant understanding of state-action transitions. Agents can then start selecting optimal actions which is essential for maximizing reward. These actions often lead the agent to spurious experiences which the agent may not have encountered. In the case of model-based RL, agents leverage spurious experiences \cite{smirl} and plan effectively for future steps. In the case of model-free RL, surprise results in sample-inefficient learning \cite{surprise}. This can be tackled by making use of rigorous exploration strategies \cite{effectiveexp,statemarginal}. However, such techniques do not necessarily scale to high-dimensional tasks and require extrinsic feature engineering \cite{hdqn} in conjunction with meta models \cite{metaexp}. A suitable way to tackle high-dimensional dynamics is by utilizing surprise as a penalty on the reward \cite{gen}. This leads to improved generalization. However, such solutions do not show evidence of improvement in multiple agents \cite{marlsurp}. 

\subsection{Overestimation Bias}
Recent advances \cite{td3} in value-based methods have addressed overestimation bias (also known as approximation error) which stems from the value estimates approximated by the function approximator. Such methods make use of dual target functions \cite{duel} which improve stability in the Bellman updates. This has led to a significant improvement in single-agent off-policy RL methods \cite{sac}. However, MARL value-based methods continue to suffer from overestimation bias \cite{marlover,iqn}. 

\begin{figure}[H]
    \centering
    \includegraphics[height=3cm,width=8cm]{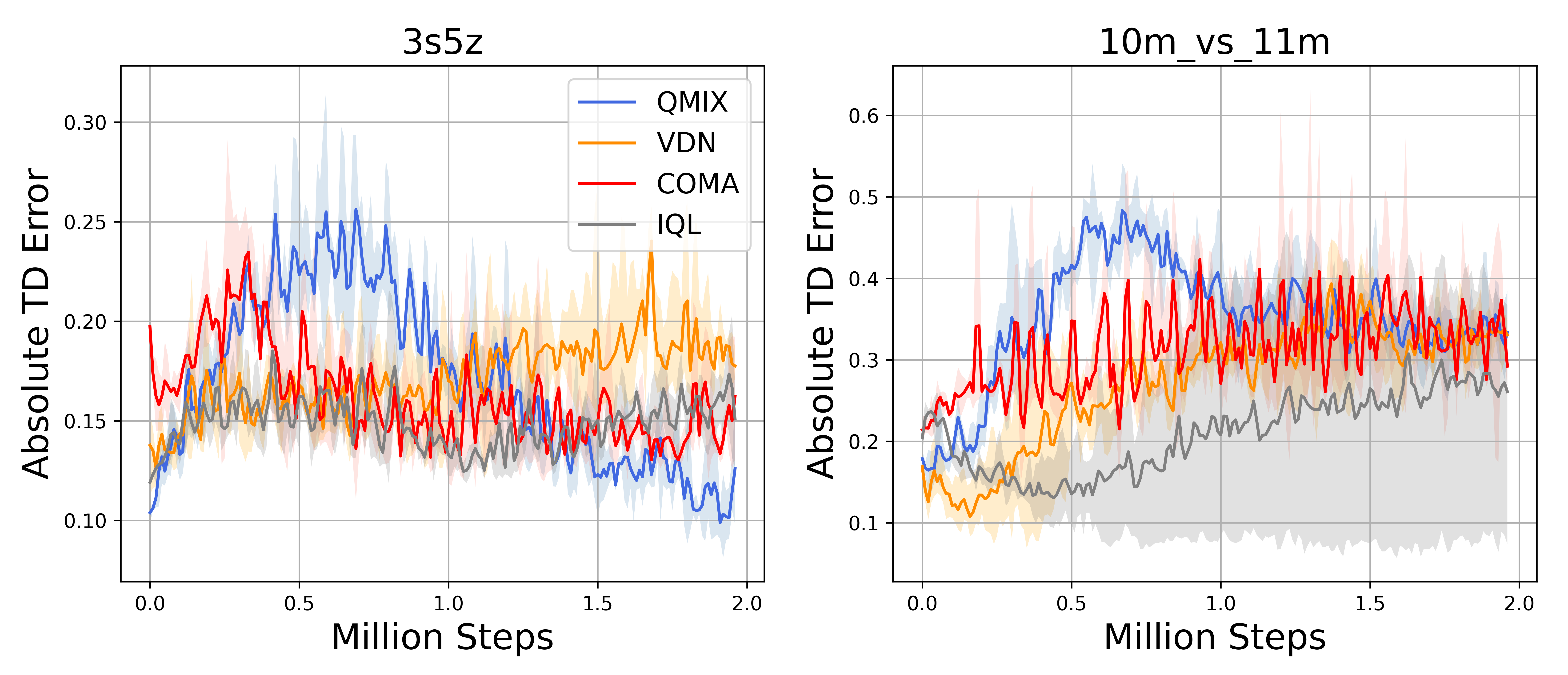}
    \caption{Absolute TD error for state-of-the-art MARL methods in StarCraft II micromanagement scenarios. Rise in error values depict the overoptimistic approximations estimated by the target value estimator.}
    \label{fig:tderror}
\end{figure}

\autoref{fig:tderror} highlights the overestimation bias originating from the overoptimistic estimations of the target value estimator. Plots present the variation of absolute TD error during learning for state-of-the-art MARL methods, namely Independent Q-Learning \cite{iql}, Counterfactual Multi-Agent Policy Gradients (COMA) \cite{coma}, VDN \cite{vdn} and QMIX \cite{qmix}. Significant rise in error values of value factorization methods such as QMIX and VDN presents the aggregation of errors from individual $Q$-value functions.

Various MARL methods \cite{twinmix} make use of a dual architecture approach which increases the stability in value factorization. Another suitable approach observed in literature is the usage of weighted bellman updates in double Q-learning \cite{weightdouble}. The Weighted Double Deep $Q$-Network (WDDQN) provides stability and sample efficiency for fully-observable MDPs. In the case of cooperative POMDPS, Weighted-QMIX (WQMIX) \cite{wqmix} yields a more sophisticated weighting scheme which aids in the retrieval of optimal policy \cite{challenges}. Although suitable for value factorization in challenging micromanagement tasks, the method needs to be carefully hand-engineered and, in the case of multiple weighting schemes, does not include a basis for selection. A more practical approach in the case of single-agent methods is the use of a family of $Q$-functions \cite{maxmin} wherein each estimator is optimized individually. Such a framework provides a generalized method for training agents with greedy policies and minimum approximation error. Although successful in single-agent settings, generalized Q-function methods do not demonstrate evidence for scalability in the number of agents \cite{challenges}.  

\subsection{Energy-based Models}
Energy-Based Models (EBMs) \cite{ebm,ebmdoc} have been successfully applied in the field of machine learning \cite{energy} and probabilistic inference \cite{david}. A typical EBM $\mathcal{E}$ formulates the equilibrium probabilities \cite{rlhinton} $P(v,h) = \frac{\exp{(-\mathcal{E}(v,h))}}{\sum_{\hat{v},\hat{h}}[\exp{(-\mathcal{E}(\hat{v},\hat{h}))}]}$ via a Boltzmann distribution \cite{boltzmann} where $v$ and $h$ are the values of the visible and hidden variables and $\hat{v}$ and $\hat{h}$ are all the possible configurations of the visible and hidden variables respectively. The probability distribution over all the visible variables can be obtained by summing over all possible configurations of the hidden variables. This is mathematically expressed in \autoref{eq:1}.
\begin{gather}
    P(v) = \frac{\sum_{h}[\exp{(-\mathcal{E}(v,h))}]}{\sum_{\hat{v},\hat{h}}[\exp{(-\mathcal{E}(\hat{v},\hat{h}))}]} \label{eq:1}    
\end{gather}
Here, $\mathcal{E}(v,h)$ is called the equilibrium free energy which is the minimum of the variational free energy and $\sum_{\hat{v},\hat{h}}[\exp{(-\mathcal{E}(\hat{v},\hat{h}))}]$ is the partition function.

\indent EBMs have been successfully implemented in single-agent RL methods \cite{pgq,sql}. These typically make use of Boltzmann distributions to approximate policies \cite{boltzmann}. Such a formulation results in the minimization of free energy within the agent. While policy approximation depicts promise in the case of unknown dynamics, inference methods \cite{inference} play a key role in optimizing goal-oriented behavior. 

A second type of usage of EBMs follows the maximization of entropy \cite{ziebartinverse}. The maximum entropy framework \cite{sac} highlighted in Soft Q-Learning (SQL) \cite{sql} allows the agent to obey a policy which maximizes its reward and entropy concurrently. Maximization of agent's entropy results in diverse and adaptive behaviors \cite{ziebartphd}.. Moreover, the maximum entropy framework is equivalent to approximate inference in the case of policy gradient methods \cite{equivalence}. Such a connection between likelihood ratio gradient techniques and energy-based formulations leads to diverse and robust policies \cite{haarnoja} and their hierarchical extensions \cite{hierarchical} which preserve the lower levels of hierarchies.

In the case of MARL, EBMs have witnessed limited applicability as a result of the increasing number of agents and complexity within each agent \cite{overview}. While the probabilistic framework is readily transferable to opponent-aware multi-agent systems \cite{probabilistic}, cooperative settings consisting of coordination between agents require a firm formulation of energy which is scalable in the number of agents \cite{twoplayer} and accounts for environments consisting of spurious states \cite{marlsql}. 

\section{Energy-based Surprise Minimization}
In this section we introduce the novel surprise minimizing EMIX agent. The motivation behind EMIX stems from spurious states and overestimation bias among agents in the case of partially-observed settings. EMIX aims to address these challenges by making use of an energy-based surprise value function in conjunction with dual target function approximators. 

\subsection{The Surprise Minimization Objective}
Firstly, we formulate the energy-based objective consisting of surprise as a function of states $s$, joint actions $u$ and standard deviations $\sigma$ within states for each agent $a$. We call this function as the surprise value function $V_{surp}^{a}(s,u,\sigma)$ which serves as a mapping from agent and environment dynamics to surprise. We then define an energy operator presented in \autoref{eq:2} which sums the free energy across all agents. 
\begin{gather}
    \mathcal{T}V^{a}_{surp}(s,u,\sigma) = \log \sum_{a=1}^{N} \exp{(V^{a}_{surp}(s,u,\sigma))} \label{eq:2}
\end{gather}
We make use of the Mellowmax operator \cite{mellowmax} as our energy operator. The energy operator is similar to the SQL energy formulation \cite{sql} where the energy across different actions is evaluated. In our case, inference is carried out across all agents with actions as prior variables. However, in the special case of using an EBM as a $Q$-function, the EMIX objective reduces to the SQL objective. We direct the curious reader to \autoref{sc:connection} for details on connection between SQL and our energy formulation.

Our choice of the energy operator is based on its unique mathematical properties which result in better convergence. Of these properties, the most useful result is that the energy operator forms a contraction on the surprise value function indicating a guaranteed minimization of surprise within agents. This is formally stated in Theorem \autoref{one}. We defer proof of Theorem \autoref{one} to \autoref{sc:proofs}.
\begin{customthm}{1}\label{one}
Given a surprise value function $V^{a}_{surp}(s,u,\sigma) \forall a \in N$, the energy operator $\mathcal{T}V^{a}_{surp}(s,u,\sigma)=\log \sum_{a=1}^{N} \exp{(V^{a}_{surp}(s,u,\sigma))}$ forms a contraction on $V^{a}_{surp}(s,u,\sigma)$. 
\end{customthm}
The energy-based surprise minimization objective can then be formulated by simply adding the approximated energy-based surprise to the initial Bellman objective as expressed below.
\begin{multline*}
    L(\theta) = \underset{b \sim R}{\mathbb{E}}[\frac{1}{2}(y - (Q(u,s;\theta)\\ + \beta \log \sum_{a=1}^{N} \exp{(V^{a}_{surp}(s,u,\sigma))}))^{2}] \nonumber
\end{multline*}
\begin{multline*}
     = \underset{b \sim R}{\mathbb{E}}[\frac{1}{2}(r + \gamma \underset{u^{'}}{\max}Q(u^{'},s^{'};\theta^{-})\\ + \beta \log \sum_{a=1}^{N} \exp{(V^{a}_{surp}(s^{'},u^{'},\sigma^{'}))}\\ - (Q(u,s;\theta) + \beta \log \sum_{a=1}^{N} \exp{(V^{a}_{surp}(s,u,\sigma))}))^{2}]
\end{multline*}
\begin{multline*}
     = \underset{b \sim R}{\mathbb{E}}[\frac{1}{2}(r + \gamma \underset{u^{'}}{\max}Q(u^{'},s^{'};\theta^{-})\\ + \beta \log \frac{\sum_{a=1}^{N} \exp{(V^{a}_{surp}(s^{'},u^{'},\sigma^{'}))}}{\sum_{a=1}^{N} \exp{(V^{a}_{surp}(s,u,\sigma))}} - Q(u,s;\theta))^{2}]
\end{multline*}
\begin{multline*}
    L(\theta) = \underset{b \sim R}{\mathbb{E}}[\frac{1}{2}(r + \gamma \underset{u^{'}}{\max}Q(u^{'},s^{'};\theta^{-})\\ + \beta E - Q(u,s;\theta))^{2}]
\end{multline*}
Here, $E$ is defined as the surprise ratio with $\beta$ as a temperature parameter and $\sigma^{'}$ as the deviation among next states in the batch. The surprise value function is approximated by a universal function approximator (in our case a neural network) with its parameters as $\phi$. $V_{a}(s^{'},u^{'},\sigma^{'})$ is expressed as the negative free energy and $\sum_{a=1}^{N} \exp{(V_{a}(s,u,\sigma))}$ the partition function. Alternatively, $V_{a}(s,u,\sigma)$ can be formulated as the negative free energy with $\sum_{a=1}^{N} \exp{(V_{a}(s^{'},u^{'},\sigma^{'}))}$ as the partition function. The objective incorporates the minimization of surprise across all agents as minimizing the energy in \textit{surprising} states. Such a formulation of surprise acts as intrinsic motivation and at the same time provides robustness to multi-agent behavior. Furthermore, the energy formulation in the form of energy ratio $E$ is a suitable one as it guarantees convergence to minimum surprise at optimal policy $\pi^{*}$. This is formally expressed in Theorem \autoref{two} with its corresponding proof in \autoref{sc:proofs}.
\begin{customthm}{2}\label{two}
Upon agent's convergence to an optimal policy $\pi^{*}$, total energy of $\pi^{*}$, expressed by $E^{*}$ will reach a thermal equilibrium consisting of minimum surprise among consecutive states $s$ and $s^{'}$.
\end{customthm}
The objective can be modified to address approximation error in the target $Q$-values using Maxmin Q-learning \cite{maxmin}. We introduce a total of $m$ target approximators making $\{Q_{1}(u^{'},s^{'};\theta^{-}),Q_{2}(u^{'},s^{'};\theta^{-})...,Q_{m}(u^{'},s^{'};\theta^{-})\}$ as the set of target approximators. This allows the objective to address overestimation bias in a scalable manner without using multiple $Q$-functions at the same time. The final EMIX objective is mathematically expressed in \autoref{eq:4}.
\begin{multline*}
    L(\theta) = \underset{b \sim R}{\mathbb{E}}[\frac{1}{2}(r + \gamma \underset{u^{'}}{\max}\underset{i}{\min}Q_{i}(u^{'},s^{'};\theta^{-})\\ + \beta E - Q(u,s;\theta))^{2}] \label{eq:4}
\end{multline*}

Here, $i$ depicts each of the $m$ target estimators with $\underset{i}{\min}Q_{i}(u^{'},s^{'};\theta^{-})$ indicating the estimate with minimum error. 

\subsection{Energy-based MIXer (EMIX)}
\begin{algorithm}[H]
\caption{Energy-based MIXer (EMIX)}
\label{alg:algorithm1}
\begin{algorithmic}[1]

  \State Initialize $\phi$, $\theta$, $\theta_{1}^{-}...,\theta_{m}^{-}$, agent and hypernetwork parameters.
  \State Initialize learning rate $\alpha$, temperature $\beta$ and replay buffer $\mathcal{R}$.

  \For{environment step}
      \State $u \xleftarrow[]{} (u_{1},u_{2}...,u_{N})$
      \State $\mathcal{R} \xleftarrow[]{} \mathcal{R} \cup \{(s,u,r,s^{'})\}$
      \If{$|\mathcal{R}| >$ batch-size}
        \For{random batch}
            \State $Q_{tot}^{\theta} \xleftarrow[]{}$ \textit{Mixer-Network}($Q_{1},Q_{2}...,Q_{N},s$)
            \State $Q_{i}^{\theta^{-}} \xleftarrow[]{}$ \textit{Target-Mixer}$_{i}$($Q_{1},Q_{2}...,Q_{N},s^{'}$),  $\forall i=1,2..,m$
            \State Calculate $\sigma$ and $\sigma^{'}$ using $s$ and $s^{'}$
            \State $V_{surp}^{a}(s,u,\sigma) \xleftarrow[]{}$ \textit{Surprise-Mixer($s,u,\sigma$)}
            \State $V_{surp}^{a}(s^{'},u^{'},\sigma^{'}) \xleftarrow[]{}$ \textit{Target-Surprise-Mixer($s^{'},u^{'},\sigma^{'}$)}
            \State $E \xleftarrow[]{} \log \frac{\sum_{a=1}^{N} \exp{(V^{a}_{surp}(s^{'},u^{'},\sigma^{'}))}}{\sum_{a=1}^{N} \exp{(V^{a}_{surp}(s,u,\sigma))}}$
            \State Calculate $L(\theta)$ using $E$ in \autoref{eq:4}
            \State $\theta \xleftarrow[]{} \theta - \alpha \nabla_{\theta}L(\theta)$
        \EndFor
    \EndIf
    \If{update-interval steps have passed}
        \State $\theta_{i}^{-} \xleftarrow[]{} \theta, \forall i=1,2..,m$
    \EndIf
  \EndFor

\end{algorithmic}
\end{algorithm}

Algorithm \autoref{alg:algorithm1} presents the EMIX algorithm. We initialize surprise value function parameters $\phi$, mixer parameters $\theta$, target parameters $\theta_{i}^{-}$ for $i=1,2...,m$ and lastly the agent and hypernetwork parameters of QMIX. A learning rate $\alpha$, temperature $\beta$ and replay buffer $\mathcal{R}$ are instantiated. During environment interactions, agents in state $s$ perform joint action $u$, observe reward $r$ and transition to next-states $s^{'}$. These experiences are stored in relay buffer $\mathcal{R}$ as $(s,u,r,s^{'})$ tuples. In order to make agents explore the environment, an $\epsilon$-greedy schedule is used similar to the original QMIX \cite{qmix} implementation. During the update steps, a random batch of $batch-size$ is sampled from $\mathcal{R}$. The total $Q$-value $Q_{tot}^{\theta}$ is computed by the mixer network with its inputs as the $Q$-values of all the agents conditioned on $s$ via the hypernetworks. Similarly, the target mixers approximate $Q_{i}^{\theta^{-}}$ conditioned on $s^{'}$. In order to evaluate surprise within agents, we compute the standard deviations $\sigma$ and $\sigma^{'}$ across all observations $z$ and $z^{'}$ for each agent using $s$ and $s^{'}$ respectively. The surprise value function called the \textit{Surprise-Mixer} estimates surprise $V^{a}_{surp}(s,u,\sigma)$ conditioned on $s$, $u$ and $\sigma$. The same computation is repeated using the Target-Surprise-Mixer for estimating surprise $V^{a}_{surp}(s^{'},u^{'},\sigma^{'})$  within next-states in the batch. Application of the energy operator for $V^{a}_{surp}(s,u,\sigma)$ and $V^{a}_{surp}(s^{'},u^{'},\sigma^{'})$ yields the energy ratio $E$ which is used in \autoref{eq:4} to evaluate $L(\theta)$. We then use batch gradient descent to update parameters of the mixer $\theta$. Target parameters $\theta_{i}^{-}$ are updated every $update-interval$ steps.

We now take a closer look at the surprise-mixer approximating the surprise value function. In order to condition surprise on states, joint actions and the deviation among states, we construct an expressive architecture motivated by provable exploration in RL \cite{homer}. Such models have proven to be efficient in the case of provable exploration \cite{homer} as it allows the agent to learn an exploration policy for every value of abstract state related to the latent space. We borrow from this technique of provable exploration and extend it to the surprise minimization setting.
\begin{figure}[H]
    \centering
    \includegraphics[height=5cm,width=6.5cm]{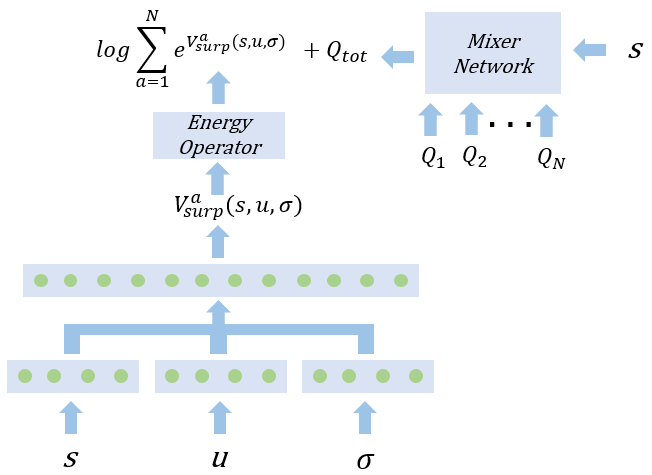}
    \caption{Surprise-Mixer architecture for estimation of the surprise value function.}
    \label{fig:emix}
\end{figure}
\autoref{fig:emix} presents the expressive architecture of surprise-mixer network utilized for surprise value function approximation and minimization. The surprise-mixer maps transitions consisting of states $s$, joint actions $u$ and deviations $\sigma$ to a surprise value $V_{surp}^{a}(s,u,\sigma)$ for all agents $a$. The architecture allows the agent to learn a robust and surprise-agnostic policy for every value of abstract state related to the latent space. Moreover, the latent space accommodates every value of surprise across agents as a result of state standard deviations induced in the intermediate representations. Surprise value estimates $V_{surp}^{a}(s,u,\sigma)$ are evaluated by the energy operator with the resulting expression becoming a part of the Bellman objective in \autoref{eq:4} comprising of the total $Q$-values $Q_{tot}$ estimated by the mixer network.


\section{Experiments}

\begin{table*}[ht]
    \centering
     \begin{tabular}{c c c c c c c} 
     \hline
     Scenarios & EMIX & SMiRL-QMIX & QMIX & VDN & COMA & IQL \\ [0.5ex] 
     \hline
     2s\textunderscore vs\textunderscore 1sc & 90.33 $\pm$ 0.72 & 88.41 $\pm$ 1.31 & 89.19 $\pm$ 3.23 & 91.42 $\pm$ 1.23 & \textbf{96.90 $\pm$ 0.54} & 86.07 $\pm$ 0.98\\ 
     2s3z & \textbf{95.40$\pm$0.45} & 94.93$\pm$0.32 & 95.30$\pm$1.28 & 92.03$\pm$2.08 & 43.33$\pm$2.70 & 55.74$\pm$6.84\\ 
     3m & \textbf{94.90$\pm$0.39} & 93.94$\pm$0.22 & 93.43$\pm$0.20 & 94.58$\pm$0.58 & 84.75$\pm$7.93 & 94.79$\pm$0.50\\ 
     3s\textunderscore vs\textunderscore 3z & \textbf{99.58$\pm$0.07} & 97.63$\pm$1.08 & 99.43$\pm$0.20 & 97.90$\pm$0.58 & 0.21$\pm$0.54 & 92.32$\pm$2.83\\ 
     3s\textunderscore vs\textunderscore 4z & \textbf{97.22$\pm$0.73} & 0.24$\pm$0.11 & 96.01$\pm$3.93 & 94.29$\pm$2.13 & 0.00$\pm$0.00 & 59.75$\pm$12.22\\ 
     3s\textunderscore vs\textunderscore 5z & 52.91$\pm$11.80 & 0.00$\pm$0.00 & 43.44$\pm$7.09 & \textbf{68.51$\pm$5.60} & 0.00$\pm$0.00 & 18.14$\pm$2.34\\ 
     3s5z & \textbf{88.88$\pm$1.07} & 88.53$\pm$1.03 & 88.49$\pm$2.32 & 63.58$\pm$3.99 & 0.25$\pm$0.11 & 7.05$\pm$3.52\\ 
     8m & \textbf{94.47$\pm$1.38} & 89.96$\pm$1.42 & 94.30$\pm$2.90 & 90.26$\pm$1.12 & 92.82$\pm$0.53 & 83.53$\pm$1.62\\ 
     8m\textunderscore vs\textunderscore 9m & \textbf{71.03$\pm$2.69} & 69.90$\pm$1.94 & 68.28$\pm$2.30 & 58.81$\pm$4.68 & 4.17$\pm$0.58 & 28.48$\pm$22.38\\ 
     10m\textunderscore vs\textunderscore 11m & 75.35$\pm$2.30 & \textbf{77.85$\pm$2.02} & 70.36$\pm$2.87 & 71.81$\pm$6.50 & 4.55$\pm$0.73 & 32.27$\pm$25.68\\ 
     so\textunderscore many\textunderscore baneling & \textbf{95.87$\pm$0.16} & 93.61$\pm$0.94 & 93.35$\pm$0.78 & 92.26$\pm$1.06 & 91.65$\pm$2.26 & 74.97$\pm$6.52\\ 
     5m\textunderscore vs\textunderscore 6m & \textbf{37.07$\pm$2.42} & 33.27$\pm$2.79 & 34.42$\pm$2.63 & 35.63$\pm$3.32 & 0.52$\pm$0.13 & 14.78$\pm$2.72\\ 
     \hline
     \end{tabular}
     \caption{Comparison of success rate percentages between EMIX and state-of-the-art MARL methods for StarCraft II micromanagement scenarios. Results are averaged over 5 random seeds with each session consisting of 2 million environment interactions. EMIX significantly improves the performance of the QMIX agent on a total of 9 out of 12 scenarios. EMIX demonstrates improved performance for surprise minimization on all 12 scenarios in comparison to the SMiRL scheme. In addition, EMIX presents less deviation between its random seeds indicating consistency in collaboration across agents.}
    \label{tab:table}
\end{table*}
    
Our experiments aim to evaluate the performance, consistency, sample-efficiency and effectiveness of the various components of our method. Specifically, we aim to answer the following questions- (1) How does our method compare to current state-of-the-art MARL methods in terms of performance, consistency and sample efficiency?, (2) How much does each component of the method contribute to its performance? and (3) Does the algorithm validate the theoretical claims corresponding to its components?

\subsection{Energy-based Surprise Minimization}
We assess the performance and sample-efficiency of EMIX on multi-agent StarCraft II micromanagement scenarios \cite{smac}. We select StarCraft II scenarios particularly for three reasons. Firstly, micromanagement scenarios consist of a larger number of agents with different action spaces. This requires a greater deal of coordination in comparison to other benchmarks \cite{predator} which attend to other aspects of MARL performance such as opponent-awareness \cite{survey}. Secondly, micromanagement scenarios consist of partial observability wherein agents are restricted from responding to enemy fire and attacking enemies when they are in range \cite{qmix}. This allows agents to explore the environment effectively and find an optimal strategy purely based on collaboration rather than built-in game utilities. Lastly, micromanagement scenarios in StarCraft II consist of multiple opponents which introduce a greater degree of surprise within consecutive states. Irrespective of the time evolution of an episode, environment dynamics of each scenario change rapidly as the agents need to respond to enemy's behavior. 

We compare our method to current state-of-the-art methods, namely QMIX \cite{qmix}, VDN \cite{vdn}, COMA \cite{coma} and IQL \cite{iql}. In order to compare our surprise-based scheme against pre-existing surprise minimization mechanisms, we compare EMIX additionally to a model-free implementation of SMiRL \cite{smirl} in QMIX. All methods were implemented using the PyMARL framework \cite{smac}. The SMiRL component was additionally incorporated as per the update rule provided in \cite{gen}. We term this implementation as SMiRL-QMIX for our comparisons. Agents were trained for a total of 5 random seeds consisting of 2 million steps in each environment. We use an $\epsilon$-greedy exploration scheme wherein $\epsilon$ is annealed from 1 to 0.01 during the initial stages of training. Details related to the implementation of EMIX are presented in \autoref{sc:implement}.

In order to assess the performance and sample-efficiency of agents we evaluate the success rate percentages of each multi-agent system in completing each scenario. A completion of a scenario indicates the victory of the team over its enemies. Scenarios consist of varying difficulties in terms of the number of agents, map locations, distance from enemies, number of enemies and the level of difficulty.
\autoref{tab:table} presents the comparison of success rate percentages between EMIX and state-of-the-art MARL algorithms on StarCraft II micromanagement scenarios. Out of the 12 scenarios considered, EMIX presents higher success rates on 9 of these scenarios depicting the suitability of the proposed approach. In scenarios such as \textit{3m}, \textit{3s5z} and \textit{8m} performance gain between EMIX and other methods such as QMIX and VDN are incremental as a result of the small number of agents and simplicity of tasks. On the other hand, EMIX presents significant performance gains in cases of \textit{so\textunderscore many\textunderscore baneling} and \textit{5m\textunderscore vs \textunderscore 6m} which consist of a large number of opponents and a greater difficulty level respectively. Complete results for all scenarios including plots presenting agents' learning performance can be viewed in \autoref{sc:results}.

When compared to QMIX, EMIX depicts improved success rates on all of the 12 scenarios. For instance, in scenarios such as \textit{3s\textunderscore vs\textunderscore 5z}, \textit{8m\textunderscore vs\textunderscore 9m} and \textit{5m\textunderscore vs\textunderscore 6m} QMIX presents sub-optimal performance. On the other hand, EMIX utilizes a comparatively improved joint policy and yields better convergence in a sample-efficient manner. Moreover, on comparing EMIX with SMiRL-QMIX, we note that EMIX demonstrates a higher average success rate. This highlights the suitability of the energy-based scheme in the case of a larger number of agents and complex environment dynamics for surprise minimization.

In addition to improved performance and sample-efficiency, EMIX also presents consistency in its learning across different random seeds. Deviation in success rates for EMIX is comparable to pre-existing value factorization methods such as QMIX and VDN. This indicates that the energy-based formulation of surprise minimization is compatible with value factorization methods and enables all the agents to exhibit the same optimal behavior across different runs. 

\subsection{Ablation Study}
\begin{figure*}[ht]
    \centering
    \includegraphics[height=3cm,width=17.5cm]{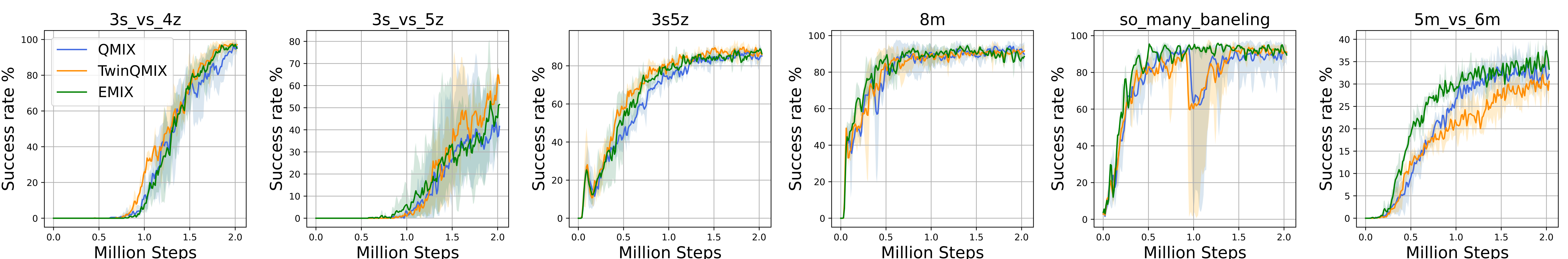}
    \includegraphics[height=3cm,width=17.5cm]{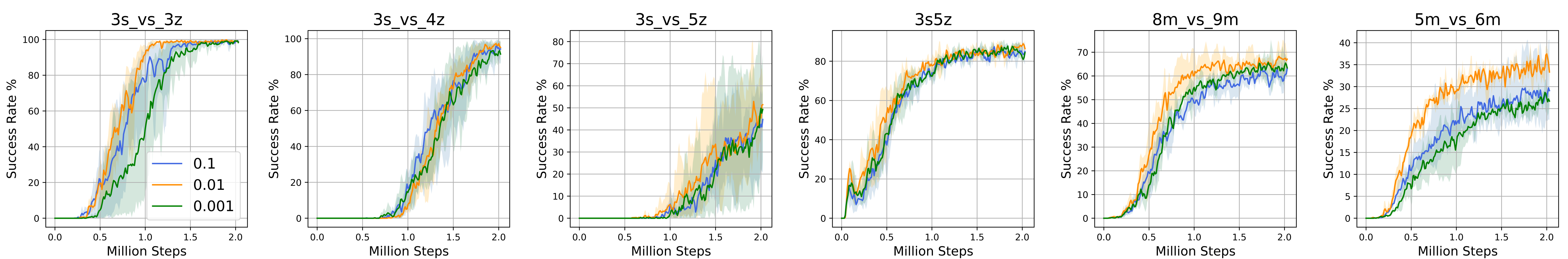}
    \includegraphics[height=3cm,width=17.5cm]{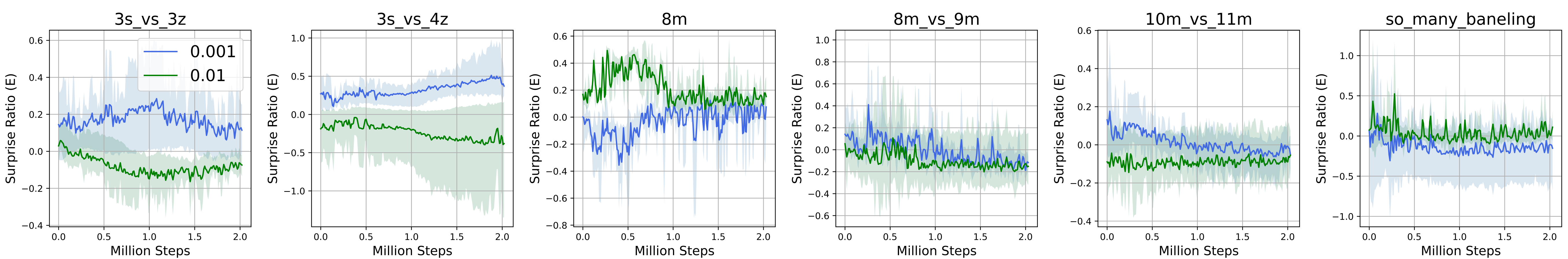}
    \caption{Ablations on six different scenarios for each of EMIX's components (top), its variation in performance (middle) and surprise minimization (bottom) with temperature $\beta$. When compared to QMIX, EMIX and TwinQMIX (EMIX without surprise minimization) depict improved performance and sample efficiency indicating the suitability of energy-based surprise minimization and dual Q-function scheme. This is achieved by making use of a suitable value of temperature parameter ($\beta=0.01$) which controls the extent of energy ratio $E$ in the EMIX objective. Moreover, the temperature parameter affects the stability in surprise minimization by utilizing $E$ as intrinsic motivation.}
    \label{fig:ablations}
\end{figure*}

We now present the ablation study for the various components of EMIX. Our experiments aim to determine the effectiveness of the energy-based surprise minimization method and the multiple target $Q$-function scheme. Additionally, we also aim to determine the extent up to which our proposed framework is viable in the standard QMIX objective.

\subsubsection{Energy-based Surprise Minimization and Overestimation Bias}
To weigh the effectiveness of the multiple target $Q$-function scheme we remove the energy-based surprise minimization from EMIX and replace it with the prior QMIX objective. For simplicity, we make use of two target $Q$-functions. We call this implementation of QMIX combined with the dual target function scheme as \textit{TwinQMIX}. Thus, we can compare between QMIX, TwinQMIX and EMIX to assess the contributions of each of the proposed methods. \autoref{fig:ablations} (top) presents the comparison of average success rates for QMIX, TwinQMIX and EMIX on 6 different scenarios with lines indicating average success rates and the shaded area as standard deviation across 5 random seeds.

In comparison to QMIX, TwinQMIX adds stability to the original objective and yields performance gains in the form of improved success rates and sample-efficient convergence. For instance, in the \textit{3s\textunderscore vs\textunderscore 5z} scenario, TwinQMIX significantly improves the performance of QMIX by reducing the overoptimistic estimates in the initial QMIX objective. However, in the \textit{5m\textunderscore vs\textunderscore 6m} scenario, TwinQMIX falls short of optimal sample efficiency as a result of underoptimistic estimates yielded by the $\underset{i}{min}Q_{i}^{-}(s,u,\sigma)$ operation.

On comparing TwinQMIX to EMIX we note that the energy-based surprise minimization scheme provides significant performance improvement in the modified QMIX objective. The EMIX objective demonstrates sample-efficiency and improved success rate values when compared to the TwinQMIX implementation. Additionally, the surprise minimization term $\beta E$ adds to the stability of the TwinQMIX objective. This is demonstrated in the \textit{5m\textunderscore vs\textunderscore 6m} scenario wherein the EMIX implementation improves the performance of TwinQMIX in comparison to QMIX by compensating for the underoptimistic estimations in the bellman updates. In the case of \textit{so\textunderscore many \textunderscore baneling} scenario, EMIX tackles surprise effectively by preventing a significant drop in performance which is observed in cases of QMIX and TwinQMIX. \textit{so\textunderscore many \textunderscore baneling} scenario consists of a large number of opponents (27 banelings) which force the agents to act quickly. This inherently induces a large amount of surprise in the form of state-to-state deviations. EMIX successfully tackles this hindrance and prevents the drop in success rates.

\subsubsection{Temperature Parameter}
We now evaluate the extent of effectiveness of our surprise minimization objective in accordance with the temperature parameter $\beta$. \autoref{fig:ablations} (middle) presents the variation of success rates of the EMIX objective with $\beta$ during learning. EMIX was evaluated for three different values (as presented in the legend) of $\beta$. While the objective is robust to significant changes in the value of $\beta$, it presents sub-optimal performance in the case of high ($\beta=0.1$) and low ($\beta=0.001$) temperature values. In the case of high $\beta$ values, the objective suffers from overestimation error in the bellman updates introduced by the energy term. The error compensates for the bias removed by the dual $Q$-function scheme. On the other hand, low $\beta$ values do not include surprise minimization and EMIX agents face surprising states.

The importance of $\beta$ can be validated by assessing its usage in surprise minimization. However, it is difficult to evaluate surprise minimization directly as surprise value function estimates $V_{surp}^{a}(s,u,\sigma)$ vary from state-to-state across different agents and present high variance during agent's learning. We instead observe the variation of energy ratio $E$ as it is a collection of surprise-based sample estimates across the batch. Additionally, $E$ consists of prior samples $V_{surp}^{a}(s,u,\sigma)$ for $V_{surp}^{a}(s^{'},u^{'},\sigma^{'})$ which makes inference across different agents tractable. \autoref{fig:ablations} (bottom) presents the variation of energy ratio $E$ with the temperature parameter $\beta$ during learning. We compare two stable variations of E at $\beta=0.001$ and $\beta=0.01$. The objective minimizes $E$ over the course of learning and attains thermal equilibrium with minimum energy. Intuitively, equilibrium corresponds to convergence to optimal policy $\pi^{*}$ which validates the claim in Theorem \autoref{two}. With $\beta=0.01$, EMIX presents improved convergence and surprise minimization for 5 out of the 6 considered scenarios, hence validating the suitable choice of $\beta$. On the other hand, a lower value of $\beta=0.001$ does little to minimize surprise across agents.


\section{Discussion}
In this paper, we introduced the Energy-based MIXer (EMIX), a multi-agent value factorization algorithm based on QMIX which minimizes surprise utilizing the energy across agents. Our method proposes a novel energy-based surprise minimization objective consisting of an energy operator in conjunction with the surprise value function. The EMIX objective satisfies theoretical guarantees of total energy and surprise minimization with experimental results validating these claims. Additionally, EMIX extends Maxmin Q-learning for addressing overestimation bias across agents in MARL. On a total 9 out of 12 challenging StarCraft II micromanagement scenarios, EMIX demonstrates improved consistent and stable performance for multi-agent surprise minimization. Ablations carried out on the proposed energy-based scheme, multiple target approximators and temperature parameter highlight the suitability of EMIX components. 

EMIX serves as a practical example of energy-based models in cooperative MARL. Surprise minimization  can be further extended towards opponent-aware and hierarchical MARL wherein agents deal with a greater degree of stochasticity under fast-paced dynamics. This extension of energy-based models would aid in gaining understanding of MARL in practical settings such as safety control and sensitivity analysis. We leave this for future work.



\section*{Acknowledgments}
We would like to thank the anonymous reviewers for providing valuable feedback on our work. We acknowledge Aravind Varier and Shashank Saurav for helpful discussions and the computing platform provided by the Department of Computer Science (DCS), University of Toronto. This work is supported by RBC Capital Markets, RBC Innovation Lab and the Center for Management of Technology and Entrepreneurship (CMTE).



\bibliography{example_paper}

\begin{thebibliography}{65}
\providecommand{\natexlab}[1]{#1}
\providecommand{\url}[1]{\texttt{#1}}
\expandafter\ifx\csname urlstyle\endcsname\relax
  \providecommand{\doi}[1]{doi: #1}\else
  \providecommand{\doi}{doi: \begingroup \urlstyle{rm}\Url}\fi

\bibitem[Achiam \& Sastry(2017)Achiam and Sastry]{surprise}
Achiam, J. and Sastry, S.
\newblock Surprise-based intrinsic motivation for deep reinforcement learning,
  2017.

\bibitem[Ackermann et~al.(2019)Ackermann, Gabler, Osa, and Sugiyama]{marlover}
Ackermann, J., Gabler, V., Osa, T., and Sugiyama, M.
\newblock Reducing overestimation bias in multi-agent domains using double
  centralized critics.
\newblock \emph{arXiv preprint arXiv:1910.01465}, 2019.

\bibitem[Asadi \& Littman(2017)Asadi and Littman]{mellowmax}
Asadi, K. and Littman, M.~L.
\newblock An alternative softmax operator for reinforcement learning.
\newblock In \emph{International Conference on Machine Learning}, 2017.

\bibitem[Berseth et~al.(2019)Berseth, Geng, Devin, Jayaraman, Finn, and
  Levine]{smirl}
Berseth, G., Geng, D., Devin, C., Jayaraman, D., Finn, C., and Levine, S.
\newblock Smirl: Surprise minimizing rl in entropic environments.
\newblock 2019.

\bibitem[Burda et~al.(2019)Burda, Edwards, Pathak, Storkey, Darrell, and
  Efros]{curiosity}
Burda, Y., Edwards, H., Pathak, D., Storkey, A., Darrell, T., and Efros, A.~A.
\newblock Large-scale study of curiosity-driven learning.
\newblock In \emph{ICLR}, 2019.

\bibitem[Busoniu et~al.(2006)Busoniu, Babuska, and De~Schutter]{survey}
Busoniu, L., Babuska, R., and De~Schutter, B.
\newblock Multi-agent reinforcement learning: A survey.
\newblock In \emph{2006 9th International Conference on Control, Automation,
  Robotics and Vision}, 2006.

\bibitem[Bu{\c{s}}oniu et~al.(2010)Bu{\c{s}}oniu, Babu{\v{s}}ka, and
  De~Schutter]{overview}
Bu{\c{s}}oniu, L., Babu{\v{s}}ka, R., and De~Schutter, B.
\newblock Multi-agent reinforcement learning: An overview.
\newblock In \emph{Innovations in multi-agent systems and applications-1}.
  2010.

\bibitem[Chen(2020)]{gen}
Chen, J.~Z.
\newblock Reinforcement learning generalization with surprise minimization,
  2020.

\bibitem[Foerster et~al.(2017)Foerster, Farquhar, Afouras, Nardelli, and
  Whiteson]{coma}
Foerster, J., Farquhar, G., Afouras, T., Nardelli, N., and Whiteson, S.
\newblock Counterfactual multi-agent policy gradients, 2017.

\bibitem[Fu et~al.(2020)Fu, Zhao, and Zhang]{twinmix}
Fu, Z., Zhao, Q., and Zhang, W.
\newblock Reducing overestimation in value mixing for cooperative deep
  multi-agent reinforcement learning.
\newblock \emph{ICAART}, 2020.

\bibitem[Fujimoto et~al.(2018)Fujimoto, van Hoof, and Meger]{td3}
Fujimoto, S., van Hoof, H., and Meger, D.
\newblock Addressing function approximation error in actor-critic methods,
  2018.

\bibitem[Grau-Moya et~al.(2018)Grau-Moya, Leibfried, and Bou-Ammar]{twoplayer}
Grau-Moya, J., Leibfried, F., and Bou-Ammar, H.
\newblock Balancing two-player stochastic games with soft q-learning.
\newblock \emph{arXiv preprint arXiv:1802.03216}, 2018.

\bibitem[Gupta et~al.(2018)Gupta, Mendonca, Liu, Abbeel, and Levine]{metaexp}
Gupta, A., Mendonca, R., Liu, Y., Abbeel, P., and Levine, S.
\newblock Meta-reinforcement learning of structured exploration strategies.
\newblock In \emph{Advances in Neural Information Processing Systems 31}. 2018.

\bibitem[Haarnoja(2018)]{haarnoja}
Haarnoja, T.
\newblock \emph{Acquiring Diverse Robot Skills via Maximum Entropy Deep
  Reinforcement Learning}.
\newblock PhD thesis, UC Berkeley, 2018.

\bibitem[Haarnoja et~al.(2017)Haarnoja, Tang, Abbeel, and Levine]{sql}
Haarnoja, T., Tang, H., Abbeel, P., and Levine, S.
\newblock Reinforcement learning with deep energy-based policies.
\newblock \emph{arXiv preprint arXiv:1702.08165}, 2017.

\bibitem[Haarnoja et~al.(2018{\natexlab{a}})Haarnoja, Hartikainen, Abbeel, and
  Levine]{hierarchical}
Haarnoja, T., Hartikainen, K., Abbeel, P., and Levine, S.
\newblock Latent space policies for hierarchical reinforcement learning.
\newblock \emph{arXiv preprint arXiv:1804.02808}, 2018{\natexlab{a}}.

\bibitem[Haarnoja et~al.(2018{\natexlab{b}})Haarnoja, Zhou, Abbeel, and
  Levine]{sac}
Haarnoja, T., Zhou, A., Abbeel, P., and Levine, S.
\newblock Soft actor-critic: Off-policy maximum entropy deep reinforcement
  learning with a stochastic actor.
\newblock \emph{arXiv preprint arXiv:1801.01290}, 2018{\natexlab{b}}.

\bibitem[Hasselt(2010)]{doubleqlearning}
Hasselt, H.~V.
\newblock Double q-learning.
\newblock In \emph{Advances in Neural Information Processing Systems 23}. 2010.

\bibitem[Hasselt et~al.(2016)Hasselt, Guez, and Silver]{deepdoubleqlearning}
Hasselt, H.~v., Guez, A., and Silver, D.
\newblock Deep reinforcement learning with double q-learning.
\newblock In \emph{Proceedings of the Thirtieth AAAI Conference on Artificial
  Intelligence}, 2016.

\bibitem[Hessel et~al.(2017)Hessel, Modayil, Van~Hasselt, Schaul, Ostrovski,
  Dabney, Horgan, Piot, Azar, and Silver]{rainbow}
Hessel, M., Modayil, J., Van~Hasselt, H., Schaul, T., Ostrovski, G., Dabney,
  W., Horgan, D., Piot, B., Azar, M., and Silver, D.
\newblock Rainbow: Combining improvements in deep reinforcement learning.
\newblock \emph{arXiv preprint arXiv:1710.02298}, 2017.

\bibitem[Kaiser et~al.(2019)Kaiser, Babaeizadeh, Milos, Osinski, Campbell,
  Czechowski, Erhan, Finn, Kozakowski, Levine, Mohiuddin, Sepassi, Tucker, and
  Michalewski]{mbrl}
Kaiser, L., Babaeizadeh, M., Milos, P., Osinski, B., Campbell, R.~H.,
  Czechowski, K., Erhan, D., Finn, C., Kozakowski, P., Levine, S., Mohiuddin,
  A., Sepassi, R., Tucker, G., and Michalewski, H.
\newblock Model-based reinforcement learning for atari, 2019.

\bibitem[Kraemer \& Banerjee(2016)Kraemer and Banerjee]{dec}
Kraemer, L. and Banerjee, B.
\newblock Multi-agent reinforcement learning as a rehearsal for decentralized
  planning.
\newblock \emph{Neurocomputing}, 190, 02 2016.

\bibitem[Kulkarni et~al.(2016)Kulkarni, Narasimhan, Saeedi, and
  Tenenbaum]{hdqn}
Kulkarni, T.~D., Narasimhan, K., Saeedi, A., and Tenenbaum, J.
\newblock Hierarchical deep reinforcement learning: Integrating temporal
  abstraction and intrinsic motivation.
\newblock In \emph{Advances in neural information processing systems}, 2016.

\bibitem[Lan et~al.(2020)Lan, Pan, Fyshe, and White]{maxmin}
Lan, Q., Pan, Y., Fyshe, A., and White, M.
\newblock Maxmin q-learning: Controlling the estimation bias of q-learning.
\newblock In \emph{International Conference on Learning Representations}, 2020.

\bibitem[LeCun et~al.(2006)LeCun, Chopra, Hadsell, Ranzato, and Huang]{ebm}
LeCun, Y., Chopra, S., Hadsell, R., Ranzato, M., and Huang, F.
\newblock A tutorial on energy-based learning.
\newblock \emph{Predicting structured data}, 1, 2006.

\bibitem[LeCun et~al.(2007)LeCun, Chopra, Ranzato, and Huang]{ebmdoc}
LeCun, Y., Chopra, S., Ranzato, M., and Huang, F.-J.
\newblock Energy-based models in document recognition and computer vision.
\newblock In \emph{Ninth International Conference on Document Analysis and
  Recognition (ICDAR 2007)}, volume~1, 2007.

\bibitem[Lee et~al.(2019)Lee, Eysenbach, Parisotto, Xing, Levine, and
  Salakhutdinov]{statemarginal}
Lee, L., Eysenbach, B., Parisotto, E., Xing, E., Levine, S., and Salakhutdinov,
  R.
\newblock Efficient exploration via state marginal matching.
\newblock \emph{arXiv preprint arXiv:1906.05274}, 2019.

\bibitem[Levine \& Abbeel(2014)Levine and Abbeel]{boltzmann}
Levine, S. and Abbeel, P.
\newblock Learning neural network policies with guided policy search under
  unknown dynamics.
\newblock In \emph{Advances in Neural Information Processing Systems}, 2014.

\bibitem[Lillicrap et~al.(2015)Lillicrap, Hunt, Pritzel, Heess, Erez, Tassa,
  Silver, and Wierstra]{ddpg}
Lillicrap, T.~P., Hunt, J.~J., Pritzel, A., Heess, N. M.~O., Erez, T., Tassa,
  Y., Silver, D., and Wierstra, D.
\newblock Continuous control with deep reinforcement learning.
\newblock \emph{CoRR}, abs/1509.02971, 2015.

\bibitem[Lowe et~al.(2017)Lowe, Wu, Tamar, Harb, Abbeel, and Mordatch]{maddpg}
Lowe, R., Wu, Y., Tamar, A., Harb, J., Abbeel, P., and Mordatch, I.
\newblock Multi-agent actor-critic for mixed cooperative-competitive
  environments, 2017.

\bibitem[Lyu \& Amato(2020)Lyu and Amato]{iqn}
Lyu, X. and Amato, C.
\newblock Likelihood quantile networks for coordinating multi-agent
  reinforcement learning.
\newblock In \emph{Proceedings of the 19th International Conference on
  Autonomous Agents and MultiAgent Systems}, 2020.

\bibitem[Macedo \& Cardoso(2005)Macedo and Cardoso]{role}
Macedo, L. and Cardoso, A.
\newblock The role of surprise, curiosity and hunger on exploration of unknown
  environments populated with entities.
\newblock In \emph{2005 portuguese conference on artificial intelligence},
  2005.

\bibitem[Macedo et~al.(2004)Macedo, Reisezein, and Cardoso]{surpmodeling}
Macedo, L., Reisezein, R., and Cardoso, A.
\newblock Modeling forms of surprise in artificial agents: empirical and
  theoretical study of surprise functions.
\newblock In \emph{Proceedings of the Annual Meeting of the Cognitive Science
  Society}, volume~26, 2004.

\bibitem[MacKay(2002)]{david}
MacKay, D. J.~C.
\newblock \emph{Information Theory, Inference \& Learning Algorithms}.
\newblock Cambridge University Press, 2002.

\bibitem[Misra et~al.(2019)Misra, Henaff, Krishnamurthy, and Langford]{homer}
Misra, D., Henaff, M., Krishnamurthy, A., and Langford, J.
\newblock Kinematic state abstraction and provably efficient rich-observation
  reinforcement learning.
\newblock \emph{arXiv preprint arXiv:1911.05815}, 2019.

\bibitem[Mnih et~al.(2013)Mnih, Kavukcuoglu, Silver, Graves, Antonoglou,
  Wierstra, and Riedmiller]{atari}
Mnih, V., Kavukcuoglu, K., Silver, D., Graves, A., Antonoglou, I., Wierstra,
  D., and Riedmiller, M.~A.
\newblock Playing atari with deep reinforcement learning.
\newblock \emph{CoRR}, abs/1312.5602, 2013.

\bibitem[Mnih et~al.(2016)Mnih, Badia, Mirza, Graves, Lillicrap, Harley,
  Silver, and Kavukcuoglu]{a3c}
Mnih, V., Badia, A.~P., Mirza, M., Graves, A., Lillicrap, T., Harley, T.,
  Silver, D., and Kavukcuoglu, K.
\newblock Asynchronous methods for deep reinforcement learning.
\newblock In \emph{International conference on machine learning}, 2016.

\bibitem[Nash(1950)]{nash}
Nash, J.~F.
\newblock Equilibrium points in n-person games.
\newblock \emph{Proceedings of the National Academy of Sciences}, 36\penalty0
  (1), 1950.

\bibitem[Nguyen et~al.(2020)Nguyen, Nguyen, and Nahavandi]{challenges}
Nguyen, T.~T., Nguyen, N.~D., and Nahavandi, S.
\newblock Deep reinforcement learning for multiagent systems: A review of
  challenges, solutions, and applications.
\newblock \emph{IEEE transactions on cybernetics}, 2020.

\bibitem[O'Donoghue et~al.(2016)O'Donoghue, Munos, Kavukcuoglu, and Mnih]{pgq}
O'Donoghue, B., Munos, R., Kavukcuoglu, K., and Mnih, V.
\newblock Combining policy gradient and q-learning.
\newblock \emph{arXiv preprint arXiv:1611.01626}, 2016.

\bibitem[Rashid et~al.(2018)Rashid, Samvelyan, de~Witt, Farquhar, Foerster, and
  Whiteson]{qmix}
Rashid, T., Samvelyan, M., de~Witt, C.~S., Farquhar, G., Foerster, J., and
  Whiteson, S.
\newblock Qmix: Monotonic value function factorisation for deep multi-agent
  reinforcement learning.
\newblock In \emph{ICML 2018: Proceedings of the Thirty-Fifth International
  Conference on Machine Learning}, 2018.

\bibitem[Rashid et~al.(2020)Rashid, Farquhar, Peng, and Whiteson]{wqmix}
Rashid, T., Farquhar, G., Peng, B., and Whiteson, S.
\newblock Weighted qmix: Expanding monotonic value function factorisation,
  2020.

\bibitem[Ren et~al.(2005)Ren, Beard, and Atkins]{marlsurp}
Ren, W., Beard, R.~W., and Atkins, E.~M.
\newblock A survey of consensus problems in multi-agent coordination.
\newblock In \emph{Proceedings of the 2005, American Control Conference,
  2005.}, 2005.

\bibitem[Sallans \& Hinton(2004)Sallans and Hinton]{rlhinton}
Sallans, B. and Hinton, G.~E.
\newblock Reinforcement learning with factored states and actions.
\newblock \emph{Journal of Machine Learning Research}, 5, 2004.

\bibitem[Samvelyan et~al.(2019)Samvelyan, Rashid, de~Witt, Farquhar, Nardelli,
  Rudner, Hung, Torr, Foerster, and Whiteson]{smac}
Samvelyan, M., Rashid, T., de~Witt, C.~S., Farquhar, G., Nardelli, N., Rudner,
  T. G.~J., Hung, C.-M., Torr, P. H.~S., Foerster, J., and Whiteson, S.
\newblock The starcraft multi-agent challenge, 2019.

\bibitem[Schrittwieser et~al.(2019)Schrittwieser, Antonoglou, Hubert, Simonyan,
  Sifre, Schmitt, Guez, Lockhart, Hassabis, Graepel, Lillicrap, and
  Silver]{shogi}
Schrittwieser, J., Antonoglou, I., Hubert, T., Simonyan, K., Sifre, L.,
  Schmitt, S., Guez, A., Lockhart, E., Hassabis, D., Graepel, T., Lillicrap,
  T., and Silver, D.
\newblock Mastering atari, go, chess and shogi by planning with a learned
  model, 2019.

\bibitem[Schulman et~al.(2017{\natexlab{a}})Schulman, Chen, and
  Abbeel]{equivalence}
Schulman, J., Chen, X., and Abbeel, P.
\newblock Equivalence between policy gradients and soft q-learning.
\newblock \emph{arXiv preprint arXiv:1704.06440}, 2017{\natexlab{a}}.

\bibitem[Schulman et~al.(2017{\natexlab{b}})Schulman, Wolski, Dhariwal,
  Radford, and Klimov]{ppo}
Schulman, J., Wolski, F., Dhariwal, P., Radford, A., and Klimov, O.
\newblock Proximal policy optimization algorithms.
\newblock \emph{CoRR}, abs/1707.06347, 2017{\natexlab{b}}.

\bibitem[Silver et~al.(2016)Silver, Huang, Maddison, Guez, Sifre, van~den
  Driessche, Schrittwieser, Antonoglou, Panneershelvam, Lanctot, Dieleman,
  Grewe, Nham, Kalchbrenner, Sutskever, Lillicrap, Leach, Kavukcuoglu, Graepel,
  and Hassabis]{go}
Silver, D., Huang, A., Maddison, C.~J., Guez, A., Sifre, L., van~den Driessche,
  G., Schrittwieser, J., Antonoglou, I., Panneershelvam, V., Lanctot, M.,
  Dieleman, S., Grewe, D., Nham, J., Kalchbrenner, N., Sutskever, I.,
  Lillicrap, T., Leach, M., Kavukcuoglu, K., Graepel, T., and Hassabis, D.
\newblock Mastering the game of {Go} with deep neural networks and tree search.
\newblock \emph{Nature}, 529\penalty0 (7587):\penalty0 484--489, January 2016.

\bibitem[Stadie et~al.(2015)Stadie, Levine, and Abbeel]{effectiveexp}
Stadie, B.~C., Levine, S., and Abbeel, P.
\newblock Incentivizing exploration in reinforcement learning with deep
  predictive models.
\newblock \emph{arXiv preprint arXiv:1507.00814}, 2015.

\bibitem[Stone \& Veloso(2000)Stone and Veloso]{predator}
Stone, P. and Veloso, M.
\newblock Multiagent systems: A survey from a machine learning perspective.
\newblock \emph{Autonomous Robots}, 8, 2000.

\bibitem[Sunehag et~al.(2018)Sunehag, Lever, Gruslys, Czarnecki, Zambaldi,
  Jaderberg, Lanctot, Sonnerat, Leibo, Tuyls, and Graepel]{vdn}
Sunehag, P., Lever, G., Gruslys, A., Czarnecki, W.~M., Zambaldi, V., Jaderberg,
  M., Lanctot, M., Sonnerat, N., Leibo, J.~Z., Tuyls, K., and Graepel, T.
\newblock Value-decomposition networks for cooperative multi-agent learning
  based on team reward.
\newblock In \emph{Proceedings of the 17th International Conference on
  Autonomous Agents and MultiAgent Systems}, AAMAS ’18, pp.\  2085–2087,
  2018.

\bibitem[Sutton \& Barto(2018)Sutton and Barto]{rl}
Sutton, R.~S. and Barto, A.~G.
\newblock \emph{Reinforcement Learning: An Introduction}.
\newblock 2018.

\bibitem[Tan(1993)]{iql}
Tan, M.
\newblock Multi-agent reinforcement learning: Independent vs. cooperative
  agents.
\newblock In \emph{In Proceedings of the Tenth International Conference on
  Machine Learning}, 1993.

\bibitem[Teh et~al.(2003)Teh, Welling, Osindero, and Hinton]{energy}
Teh, Y.~W., Welling, M., Osindero, S., and Hinton, G.~E.
\newblock Energy-based models for sparse overcomplete representations.
\newblock \emph{Journal of Machine Learning Research}, 4, 2003.

\bibitem[Thrun(1992)]{exploration}
Thrun, S.~B.
\newblock Efficient exploration in reinforcement learning.
\newblock 1992.

\bibitem[Toussaint(2009)]{inference}
Toussaint, M.
\newblock Robot trajectory optimization using approximate inference.
\newblock In \emph{Proceedings of the 26th annual international conference on
  machine learning}, 2009.

\bibitem[Vinyals et~al.(2017)Vinyals, Ewalds, Bartunov, Georgiev, Vezhnevets,
  Yeo, Makhzani, Küttler, Agapiou, Schrittwieser, Quan, Gaffney, Petersen,
  Simonyan, Schaul, van Hasselt, Silver, Lillicrap, Calderone, Keet, Brunasso,
  Lawrence, Ekermo, Repp, and Tsing]{SC2}
Vinyals, O., Ewalds, T., Bartunov, S., Georgiev, P., Vezhnevets, A.~S., Yeo,
  M., Makhzani, A., Küttler, H., Agapiou, J., Schrittwieser, J., Quan, J.,
  Gaffney, S., Petersen, S., Simonyan, K., Schaul, T., van Hasselt, H., Silver,
  D., Lillicrap, T., Calderone, K., Keet, P., Brunasso, A., Lawrence, D.,
  Ekermo, A., Repp, J., and Tsing, R.
\newblock Starcraft ii: A new challenge for reinforcement learning, 2017.

\bibitem[Vinyals et~al.(2019)Vinyals, Babuschkin, Czarnecki, Mathieu, Dudzik,
  Chung, Choi, Powell, Ewalds, Georgiev, Oh, Horgan, Kroiss, Danihelka, Huang,
  Sifre, Cai, Agapiou, Jaderberg, and Silver]{alphastar}
Vinyals, O., Babuschkin, I., Czarnecki, W., Mathieu, M., Dudzik, A., Chung, J.,
  Choi, D., Powell, R., Ewalds, T., Georgiev, P., Oh, J., Horgan, D., Kroiss,
  M., Danihelka, I., Huang, A., Sifre, L., Cai, T., Agapiou, J., Jaderberg, M.,
  and Silver, D.
\newblock Grandmaster level in starcraft ii using multi-agent reinforcement
  learning.
\newblock \emph{Nature}, 575, 11 2019.

\bibitem[Wang et~al.(2016)Wang, Schaul, Hessel, Hasselt, Lanctot, and
  Freitas]{duel}
Wang, Z., Schaul, T., Hessel, M., Hasselt, H., Lanctot, M., and Freitas, N.
\newblock Dueling network architectures for deep reinforcement learning.
\newblock In \emph{International conference on machine learning}, 2016.

\bibitem[Wei et~al.(2018)Wei, Wicke, Freelan, and Luke]{marlsql}
Wei, E., Wicke, D., Freelan, D., and Luke, S.
\newblock Multiagent soft q-learning.
\newblock \emph{arXiv preprint arXiv:1804.09817}, 2018.

\bibitem[Wen et~al.(2019)Wen, Yang, Luo, Wang, and Pan]{probabilistic}
Wen, Y., Yang, Y., Luo, R., Wang, J., and Pan, W.
\newblock Probabilistic recursive reasoning for multi-agent reinforcement
  learning.
\newblock \emph{arXiv preprint arXiv:1901.09207}, 2019.

\bibitem[Zheng et~al.(2018)Zheng, Meng, Hao, and Zhang]{weightdouble}
Zheng, Y., Meng, Z., Hao, J., and Zhang, Z.
\newblock Weighted double deep multiagent reinforcement learning in stochastic
  cooperative environments.
\newblock In \emph{Pacific Rim international conference on artificial
  intelligence}, 2018.

\bibitem[Ziebart(2010)]{ziebartphd}
Ziebart, B.~D.
\newblock Modeling purposeful adaptive behavior with the principle of maximum
  causal entropy.
\newblock 2010.

\bibitem[Ziebart et~al.(2008)Ziebart, Maas, Bagnell, and Dey]{ziebartinverse}
Ziebart, B.~D., Maas, A.~L., Bagnell, J.~A., and Dey, A.~K.
\newblock Maximum entropy inverse reinforcement learning.
\newblock In \emph{AAAI}, 2008.

\end{thebibliography}
\bibliographystyle{icml2021}

\newpage
\appendix
\onecolumn    

\addtocounter{equation}{6}
\addtocounter{table}{2}
\addtocounter{figure}{4}

\section{Proofs}
\label{sc:proofs}


\begin{customthm}{1}\label{eight}
Given a surprise value function $V^{a}_{surp}(s,u,\sigma) \forall a \in N$, the energy operator $\mathcal{T}V^{a}_{surp}(s,u,\sigma)$ $= \log \sum_{a=1}^{N} \exp{(V^{a}_{surp}(s,u,\sigma))}$ forms a contraction on $V^{a}_{surp}(s,u,\sigma)$. 
\end{customthm}

\begin{proof}
Let us first define a norm on surprise values $||V_{1} - V_{2}|| \equiv \underset{s,u,\sigma}{max}|V_{1}(s,u,\sigma) - V_{2}(s,u,\sigma)|$. Suppose $\epsilon = ||V_{1} - V_{2}||$,
\begin{gather}
    \log \sum_{a=1}^{N}\exp{(V_{1}(s,u,\sigma))} \leq \log \sum_{a=1}^{N}\exp{(V_{2}(s,u,\sigma) + \epsilon)} \nonumber \\
    = \log \sum_{a=1}^{N}\exp{(V_{1}(s,u,\sigma))} \leq \log \exp{(\epsilon)} \sum_{a=1}^{N}\exp{(V_{2}(s,u,\sigma))} \nonumber \\
    = \log \sum_{a=1}^{N}\exp{(V_{1}(s,u,\sigma))} \leq \epsilon + \log \sum_{a=1}^{N} \exp{(V_{2}(s,u,\sigma))} \nonumber \\
    = \log \sum_{a=1}^{N} \exp{(V_{1}(s,u,\sigma))} - \log \sum_{a=1}^{N} \exp{(V_{2}(s,u,\sigma))} \leq ||V_{1} - V_{2}|| \\
    = \log \sum_{a=1}^{N} \exp{(V_{1}(s,u,\sigma))} - \log \sum_{a=1}^{N} \exp{(V_{2}(s,u,\sigma))} < \gamma ||V_{1} - V_{2}|| \quad (\text{as} \;\gamma < 1) \label{eq:case1}
\end{gather}
Similarly, using $\epsilon$ with $\log \sum_{a=1}^{N} \exp{(V_{1}(s,u,\sigma))}$,
\begin{gather}
    \log \sum_{a=1}^{N}\exp{(V_{1}(s,u,\sigma) + \epsilon)} \geq \log \sum_{a=1}^{N}\exp{(V_{2}(s,u,\sigma))} \nonumber \\
    = \log \exp{(\epsilon)}\sum_{a=1}^{N}\exp{(V_{1}(s,u,\sigma))} \geq \log \sum_{a=1}^{N}\exp{(V_{2}(s,u,\sigma))} \nonumber \\
    = \epsilon + \log \sum_{a=1}^{N} \exp{(V_{1}(s,u,\sigma))} \geq \log \sum_{a=1}^{N}\exp{(V_{2}(s,u,\sigma))} \nonumber \\
    = ||V_{1} - V_{2}|| \geq \log \sum_{a=1}^{N} \exp{(V_{2}(s,u,\sigma))} - \log \sum_{a=1}^{N} \exp{(V_{1}(s,u,\sigma))} \\ 
    \gamma ||V_{1} - V_{2}|| > \log \sum_{a=1}^{N} \exp{(V_{2}(s,u,\sigma))} - \log \sum_{a=1}^{N} \exp{(V_{1}(s,u,\sigma))} \quad (\text{as} \; \gamma < 1) \label{eq:case2}
\end{gather}
Results in \autoref{eq:case1} and \autoref{eq:case2} prove that the energy operation is a contraction. 
\end{proof}

\begin{customthm}{2}\label{ten}
\label{theorem3}
Upon agent's convergence to an optimal policy $\pi^{*}$, total energy of $\pi^{*}$, expressed by $E^{*}$ will reach a thermal equilibrium consisting of minimum surprise among consecutive states $s$ and $s^{'}$.
\end{customthm}

\begin{proof}
    We begin by initializing a set of $M$ policies $\{\pi_{1},\pi_{2}...,\pi_{M}\}$ having energy ratios $\{E_{1},E_{2}...,E_{M}\}$. Consider a policy $\pi_{1}$ with surprise value function $V_{1}$. $E_{1}$ can then be expressed as
    \begin{gather}
        E_{1} = \log [\frac{\sum_{a=1}^{N}\exp{(V_{1}^{a}(s^{'},u^{'},\sigma^{'}))}}{\sum_{a=1}^{N}\exp{(V_{1}^{a}(s,u,\sigma))}}] \nonumber
    \end{gather}
    Assuming a constant surprise between $s$ and $s^{'}$, we can express $V_{1}^{a}(s^{'},u^{'},\sigma^{'}) = V_{1}^{a}(s,u,\sigma) + \zeta_{1}$ where $\zeta_{1}$ is a constant. Using this expression in $E_{1}$ we get,
    \begin{gather}
        E_{1} = \log [\frac{\sum_{a=1}^{N}\exp{(V_{1}^{a}(s,u,\sigma) + \zeta_{1}})}{\sum_{a=1}^{N}\exp{(V_{1}^{a}(s,u,\sigma))}}] \nonumber \\
        E_{1} = \log [\frac{\exp{(\zeta_{1})}\sum_{a=1}^{N}\exp{(V_{1}^{a}(s,u,\sigma))}}{\sum_{a=1}^{N}\exp{(V_{1}^{a}(s,u,\sigma))}}] \nonumber \\    
        E_{1} = \zeta_{1} \nonumber        
    \end{gather}
    Similarly, $E_{2}=\zeta_{2}$,$E_{3}=\zeta_{3}$...,$E_{M}=\zeta_{M}$. Thus, the energy residing in policy $\pi$ is proportional to the surprise between consecutive states $s$ and $s^{'}$. Clearly, an optimal policy $\pi^{*}$ is the one with minimum surprise. Mathematically,
    \begin{gather}
        \pi^{*} \geq \pi_{1},\pi_{2}...,\pi_{M} \implies \zeta^{*} \leq \zeta_{1},\zeta_{2}...,\zeta_{M} \nonumber \\
        = \pi^{*} \geq \pi_{1},\pi_{2}...,\pi_{M} \implies E^{*} \leq E_{1},E_{2}...,E_{M} \nonumber
    \end{gather}
    Thus, proving that the optimal policy consists of minimum surprise at thermal equilibrium. 
\end{proof}

\section{Connection between EMIX and Soft Q-Learning}
\label{sc:connection}
The Soft Q-Learning objective with $V_{soft}^{\theta^{-}}(s^{'})$ and $Q_{soft}(u,s;\theta)$ as state and action value functions respectively is given by-
\begin{gather}
    J_{Q}(\theta) = \mathbb{E}_{s,u \sim R} [\frac{1}{2}(r + \gamma \mathbb{E}_{s^{'} \sim R}[V_{soft}^{\theta^{-}}(s^{'})] - Q_{soft}(u,s;\theta))^{2}] \nonumber \\
    = J_{Q}(\theta) = \mathbb{E}_{s,u \sim R} [\frac{1}{2}(r + \gamma \mathbb{E}_{s^{'} \sim R}[\log \sum_{u \in A} \exp{Q(u^{'},s^{'};\theta^{-})}] - Q_{soft}(u,s;\theta))^{2}] \nonumber
\end{gather}
The gradient of this objective can be expressed as-
\begin{gather}
    \nabla_{\theta} J_{Q}(\theta) = \mathbb{E}_{s,u \sim R} [(r + \gamma \mathbb{E}_{s^{'} \sim R}[\log \sum_{u \in A} \exp{Q(u^{'},s^{'};\theta^{-})}] - Q_{soft}(u,s;\theta))]\nabla_{\theta}Q_{soft}(u,s;\theta) \label{eq:SQLgrad}
\end{gather}

And the gradient of the EMIX objective is obtained as-  
\begin{multline*}
    L(\theta) = \mathbb{E}_{s,u,s^{'} \sim R}[\frac{1}{2}(r+ \gamma \underset{u^{'}}{\max}\underset{i}{\min}Q_{i}(u^{'},s^{'};\theta^{-}) + \beta\log(\frac{\sum_{a=1}^{N}\exp{(V^{a}_{surp}(s^{'},u^{'},\sigma^{'}))}}{\sum_{a=1}^{N}\exp{(V^{a}_{surp}(s,u,\sigma))}}) - Q(u,s;\theta)^{2})] \nonumber
\end{multline*}
\begin{multline}
    \nabla_{\theta}L(\theta) = \mathbb{E}_{s,u,s^{'} \sim R}[(r+ \gamma \underset{u^{'}}{\max}\underset{i}{\min}Q_{i}(u^{'},s^{'};\theta^{-}) + \beta\log(\frac{\sum_{a=1}^{N}\exp{(V^{a}_{surp}(s^{'},u^{'},\sigma^{'}))}}{\sum_{a=1}^{N}\exp{(V^{a}_{surp}(s,u,\sigma))}}) - Q(u,s;\theta))]\nabla_{\theta}Q(u,s;\theta)  \label{eq:EMIXgrad}
\end{multline}

Comparing \autoref{eq:SQLgrad} to \autoref{eq:EMIXgrad} we notice that Soft Q-Learning and EMIX are related to each other as they utilize energy-based models. Soft Q-Learning makes use of a discounted energy function which downweights the energy values over longer horizons. Actions consisting of lower energy configurations are given preference by making use of $Q_{soft}(u,s;\theta)$ as the negative energy. On the other hand, EMIX makes use of a constant energy function weighed by $\beta$ which minimizes surprise-based energy between consecutive states. Both the objectives can be thought of as energy minimizing models which search for an optimal energy configuration. Soft Q-Learning searches for an optimal configuration in the action space whereas EMIX favours optimal behavior on spurious states. In fact, EMIX can be realized as a special case of Soft Q-Learning if the mixer agent utilizes an energy-based policy and attains thermal equilibrium. This leads us to express Theorem \autoref{theorem2}.

\begin{customthm}{3}\label{nine}
\label{theorem2}
Given an energy-based policy $\pi_{en}$ with its target function $V(s^{'}) = \log \sum_{u \in A} \exp{Q(u^{'},s^{'};\theta^{-})}$, the surprise minimization objective $L(\theta)$ reduces to the Soft Q-Learning objective $L(\theta_{soft})$ in the special case when the variational free energy function $\sum_{a=1}^{N} \exp{(V^{a}_{surp}(s^{'},u^{'},\sigma^{'}))}$ is equal to the partition function $\sum_{a=1}^{N} \exp{(V^{a}_{surp}(s,u,\sigma))}$.
\end{customthm}

\begin{proof}
We know that the EMIX objective is given by-
\begin{multline*}
    L(\theta) = \mathbb{E}_{s,u,s^{'} \sim R}[\frac{1}{2}(r+ \gamma \underset{u^{'}}{\max}\underset{i}{\min}Q_{i}(u^{'};s^{'},\theta^{-}) + \beta\log(\frac{\sum_{a=1}^{N}\exp{(V^{a}_{surp}(s^{'},u^{'},\sigma^{'}))}}{\sum_{a=1}^{N}\exp{(V^{a}_{surp}(s,u,\sigma))}}) - Q(u,s;\theta)^{2})]
\end{multline*}
Replacing the greedy policy term $\underset{u^{'}}{\max}\underset{i}{\min}Q_{i}(u^{'},s^{'};\theta^{-})$ with the energy-based value function $V(s^{'}) = \log \sum_{u^{'} \in A} \exp{Q(u^{'},s^{'};\theta^{-})}$, we get,
\begin{multline*}
    L(\theta) = \mathbb{E}_{s,u,s^{'} \sim R}[\frac{1}{2}(r+ \gamma \mathbb{E}_{s^{'} \sim R}[V(s^{'})] + \beta\log(\frac{\sum_{a=1}^{N}\exp{(V^{a}_{surp}(s^{'},u^{'},\sigma^{'}))}}{\sum_{a=1}^{N}\exp{(V^{a}_{surp}(s,u,\sigma))}}) - Q(u,s;\theta)^{2})]
\end{multline*}
\begin{multline*}
    = L(\theta) = \mathbb{E}_{s,u,s^{'} \sim R}[\frac{1}{2}(r+ \gamma \mathbb{E}_{s^{'} \sim R}[\log \sum_{u^{'} \in A} \exp{Q(u^{'},s^{'};\theta^{-})}] + \beta\log(\frac{\sum_{a=1}^{N}\exp{(V^{a}_{surp}(s^{'},u^{'},\sigma^{'}))}}{\sum_{a=1}^{N}\exp{(V^{a}_{surp}(s,u,\sigma))}}) - Q(u,s;\theta)^{2})]
\end{multline*}
At thermal equilibrium, $\sum_{a=1}^{N} \exp{(V^{a}_{surp}(s,u,\sigma))} = \sum_{a=1}^{N} \exp{(V^{a}_{surp}(s^{'},u^{'},\sigma^{'}))}$,
\begin{multline*}
    = L(\theta) = \mathbb{E}_{s,u,s^{'} \sim R}[\frac{1}{2}(r+ \gamma \mathbb{E}_{s^{'} \sim R}[\log \sum_{u^{'} \in A} \exp{Q(u^{'},s^{'};\theta^{-})}] + \beta\log(\frac{\sum_{a=1}^{N}\exp{(V^{a}_{surp}(s^{'},u^{'},\sigma^{'}))}}{\sum_{a=1}^{N}\exp{(V^{a}_{surp}(s^{'},u^{'},\sigma^{'}))}}) - Q(u,s;\theta)^{2})]
\end{multline*}
\begin{multline*}
    = L(\theta) = \mathbb{E}_{s,u,s^{'} \sim R}[\frac{1}{2}(r+ \gamma \mathbb{E}_{s^{'} \sim R}[\log \sum_{u^{'} \in A} \exp{Q(u^{'},s^{'};\theta^{-})}] + \beta\log(1) - Q(u,s;\theta)^{2})]
\end{multline*}
\begin{multline}
    = L(\theta) = \mathbb{E}_{s,u,s^{'} \sim R}[\frac{1}{2}(r+ \gamma \mathbb{E}_{s^{'} \sim R}[\log \sum_{u^{'} \in A} \exp{Q(u^{'},s^{'};\theta^{-})}] - Q(u,s;\theta)^{2})] \label{eq:2result}
\end{multline}
\autoref{eq:2result} represents the Soft Q-Learning objective, hence proving the result. 
\end{proof}

\section{Complete Results}
\label{sc:results}

\subsection{Energy-based Surprise Minimization}
In this section we present the complete results of EMIX agents for all the 12 scenarios considered in StarCraft II micromanagement. While some scenarios depict significant performance improvements, other scenarios present incremental gains as a result of early surprise minimization during the exploration phase. 

\subsubsection{Comparison to MARL agents}
\begin{figure}[H]
    \centering
    \includegraphics[height=8cm,width=8cm]{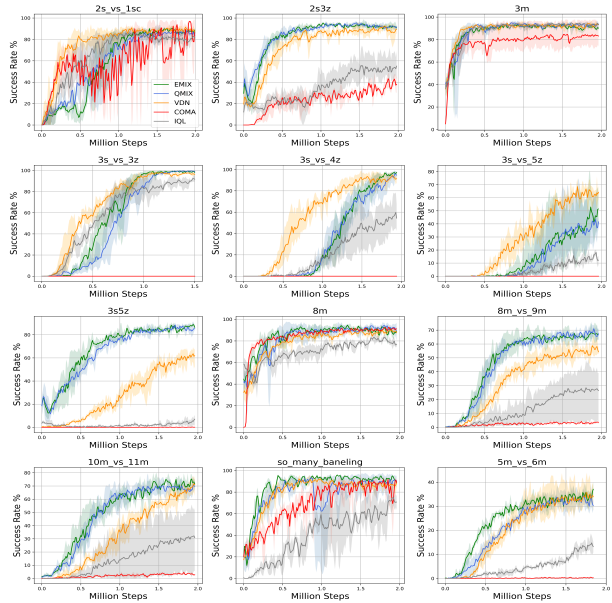}
    \caption{Learning comparison of success rate percentages between EMIX and state-of-the-art MARL methods for all StarCraft II micromanagement scenarios. Results are averaged over 5 random seeds with each session consisting of 2 million environment interactions. EMIX significantly improves the performance of the QMIX agent on a total of 9 out of 12 scenarios. In addition, EMIX presents less deviation between its random seeds indicating consistency in collaboration across agents.}
    \label{fig:rewards}
\end{figure}

\subsubsection{Comparison to SMiRL}
\begin{figure}[H]
    \centering
    \includegraphics[height=8cm,width=8cm]{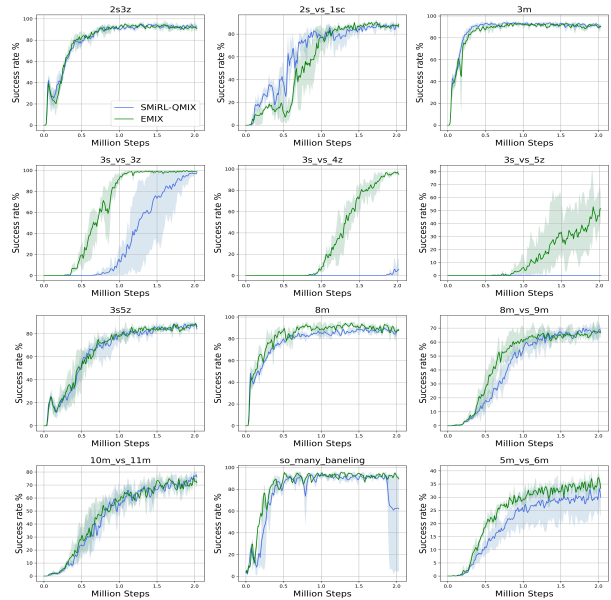}
    \caption{Learning comparison of success rate percentages between EMIX and SMiRL-QMIX for all StarCraft II micromanagement scenarios. EMIX improves the performance of QMIX for all the considered scenarios whereas the SMiRL scheme often presents sub-optimal convergence. Moreover, direct usage of standard deviations of the state distribution leads to significant approximation errors which induce sample-inefficient behavior. This can be observed from \textit{3s\textunderscore vs\textunderscore 4z} and \textit{3s\textunderscore vs\textunderscore 5z} scenarios wherein SMiRL fails to show any learning behavior. Additionally, EMIX presents less deviation between its random seeds indicating consistency in surprise minimization.}
    \label{fig:rewards}
\end{figure}

\subsection{Ablation Study}

\subsubsection{Energy-based Surprise Minimization}
\begin{figure}[H]
    \centering
    \includegraphics[height=8cm,width=8cm]{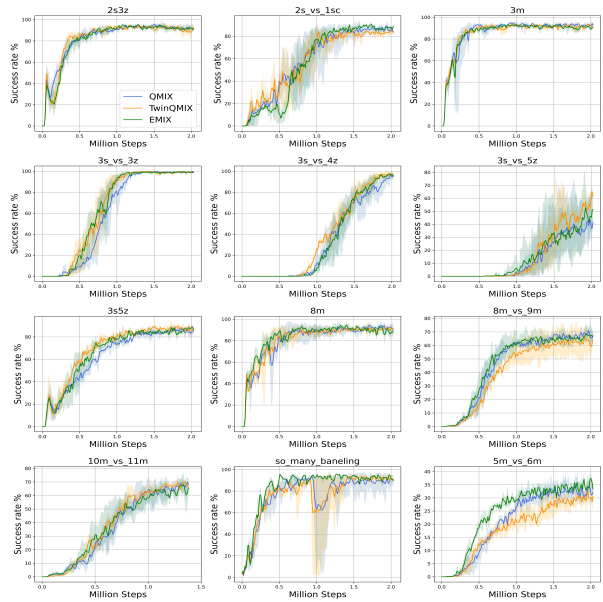}
    \caption{Comparison of success rate percentages between EMIX, TwinQMIX (EMIX without surprise minimization) and QMIX for all 12 StarCraft II micromanagement scenarios. While TwinQMIX stabilizes the performance of QMIX across agents, the surprise minimization scheme of EMIX introduces robust and sample-efficient policies.}
    \label{fig:rewards}
\end{figure}
\subsubsection{Temperature Parameter}
\begin{figure}[H]
    \centering
    \includegraphics[height=8cm,width=8cm]{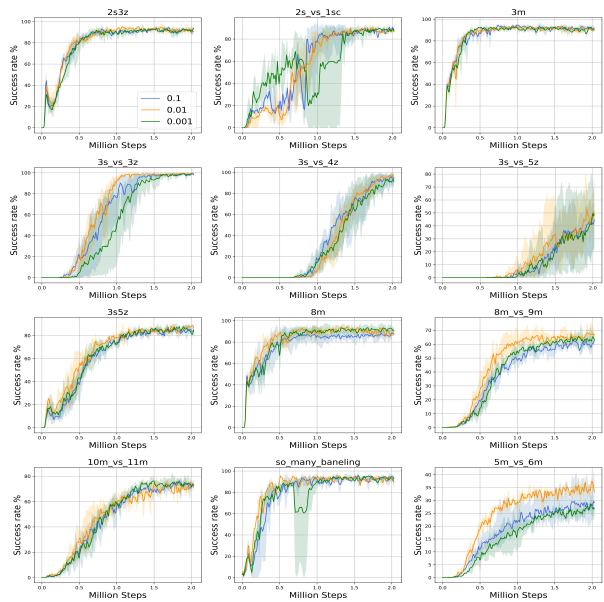}
    \caption{Comparison of success rate percentages with different $\beta$ values for EMIX on all 12 StarCraft II micromanagement scenarios. $\beta=0.01$ is a suitable value as it balances between bellman updates and the surprise minimization objective.}
    \label{fig:rewards}
\end{figure}

\section{Implementation Details}
\label{sc:implement}

\subsection{Model Specifications}
This section highlights model architecture for the surprise value function. At the lower level, the architecture consists of 3 independent networks called \textit{state\textunderscore net}, \textit{q\textunderscore net} and \textit{surp\textunderscore net}. Each of these networks consist of a single layer of 256 units with ReLU non-linearity as activations. Similar to the mixer-network, we use the ReLU non-linearity in order to provide monotonicity constraints across agents. Using a modular architecture in combination with independent networks leads to a richer extraction of joint latent transition space. Outputs from each of the networks are concatenated and are provided as input to the \textit{main\textunderscore net} consisting of 256 units with ReLU activations. The \textit{main\textunderscore net} yields a single output as the surprise value $V^{a}_{surp}(s,u,\sigma)$ which is reduced along the agent dimension by the energy operator. Alternatively, deeper versions of networks can be used in order to make the extracted embeddings increasingly expressive. However, increasing the number of layers does little in comparison to additional computational expense.  

\subsection{Hyperparameters}
\autoref{tab:hyp} presents hyperparameter values for EMIX. Value of $\beta$ was tuned between 0.001 and 1 in intervals of 0.01 with best performance observed at $\beta=0.01$. A total of 2 target $Q$-functions were used as the model is found to be robust to any greater values.  
\begin{table}[!h]
    \centering
    \begin{tabular}{c|c}
         Hyperparameters & Values \\
         \hline
         batch size & $b=32$ \\
         learning rate & $\alpha=0.0005$ \\
         discount factor & $\gamma=0.99$ \\
         target update interval & 200 episodes \\
         gradient clipping & 10 \\
         exploration schedule & $1.0$ to $0.01$ over 50000 steps\\
         mixer embedding size & 32 \\
         agent hidden size & 64 \\
         temperature & $\beta=0.01$ \\
         target $Q$-functions & 2
    \end{tabular}
    \caption{Hyperparameter values for EMIX agents}
    \label{tab:hyp}
\end{table}

\end{document}